\documentclass{article}

\usepackage{mathptmx} 
\usepackage{spconf,amsmath,graphicx}
\usepackage{tabularx}
\usepackage{cite}
\usepackage{amssymb,amsfonts}
\usepackage{algorithmic}
\usepackage{textcomp}
\usepackage{xcolor}
\usepackage{flushend}
\usepackage{enumitem}
\usepackage{booktabs}
\usepackage{multicol}
\usepackage{tikz}
\usetikzlibrary{shapes.geometric, arrows, positioning, fit}
\usepackage{multirow}
\usepackage{xurl}
\usepackage{hyperref}

\setlist{nosep, leftmargin=14pt}

\title{WBCBench 2026: A Challenge for Robust White Blood Cell Classification Under Class Imbalance}

\name{\begin{tabular}{c}
    Xin Tian$^{1}$, Xudong Ma$^{1}$, Tianqi Yang$^{2}$, Alin Achim$^{3}$, \\
    Bart\l omiej W Papie\.z$^{1}$, Phandee Watanaboonyongcharoen$^{4}$ and \textit{Nantheera Anantrasirichai}$^{3}$
  \end{tabular}
  }

\address{$^{1}$University of Oxford, Oxford, United Kingdom \\
$^{2}$University College London, London, United Kingdom \\
$^{3}$University of Bristol, Bristol, United Kingdom \\
$^{4}$Chulalongkorn University, Thailand \\
}

\begin{document}
\ninept
\maketitle

\begin{abstract}
We present \emph{WBCBench 2026}, an ISBI challenge and benchmark for automated WBC classification designed to stress-test algorithms under three key difficulties: (i) severe class imbalance across 13 morphologically fine-grained WBC classes, (ii) strict patient-level separation between training, validation and test sets, and (iii) synthetic scanner- and setting-induced domain shift via controlled noise, blur and illumination perturbations. All images are single-site microscopic blood smear acquisitions with standardised staining and expert hematopathologist annotations. This paper reviews the challenge and summarises the proposed solutions and final outcomes. The benchmark is organised into two phases. Phase~1 provides a pristine training set. Phase~2 introduces degraded images with split-specific severity distributions for train, validation and test, emulating a realistic shift between development and deployment conditions. We specify a standardised submission schema, open-source evaluator, and macro-averaged F1 score as the primary ranking metric. 

Competition websites:  \href{https://xudong-ma.github.io/WBCBench2026-Robust-White-Blood-Cell-Classification}{https://xudong-ma.github.io/WBCBench \\ 2026-Robust-White-Blood-Cell-Classification} and \\ \href{https://www.kaggle.com/competitions/wbc-bench-2026}{https://www.kaggle.com/competitions/wbc-bench-2026}.

\end{abstract}



\section{Introduction}
\label{sec:intro}

White blood cell (WBC) morphology is central to diagnosing and monitoring haematologic and immunologic disorders, including leukaemia, myelodysplastic syndromes and severe infections. In routine practice, haematologists inspect Wright--Giemsa stained peripheral blood smears to quantify and characterise WBC types such as neutrophils, lymphocytes, monocytes, eosinophils, basophils and blasts. This manual process is time-consuming, operator-dependent and sensitive to inter-observer variability and fatigue~\cite{automation_challenges_fuentes}.

Commercial digital haematology systems integrated into analysers aim to alleviate workload; however, their performance on atypical and rare classes remains limited. For example, the Sysmex DI-60 can support workflow efficiency but requires extensive expert review and shows suboptimal accuracy, particularly for abnormal cells relevant to leukaemia diagnosis~\cite{sysmex_zhao, sysmex_kim, sysmex_tabe, sysmex_nam}. Such systems are also costly and difficult to deploy at scale in low-resource settings.

Deep learning has substantially advanced WBC image classification on public datasets, often achieving near-perfect performance on balanced, pristine images. Yet these advances hide an uncomfortable reality: most models are evaluated on small cohorts with weak patient-level separation, limited blast coverage, and minimal acquisition variability. As a result, these systems remain not robust to domain shift, class imbalance, and real-world degradation.

To address this gap, we present \emph{WBCBench 2026}, an ISBI challenge that provides:
\begin{itemize}
    \item a clinically realistic, single-site dataset of blood smear images with 13 expert-annotated WBC classes, including blasts and rare subtypes;
    \item a multi-phase splitting strategy with strict patient-level separation and group-stratified sampling to preserve minority coverage;
    \item a controlled degradation pipeline that injects scanner- and setting-like variability (e.g.\ blur, noise, illumination, colour) exclusively into Phase~2;
    \item a standardised evaluation protocol based on macro-averaged F1, along with an open-source evaluator and leaderboard.
\end{itemize}

\begin{figure}
    \centering
    \includegraphics[width=\columnwidth]{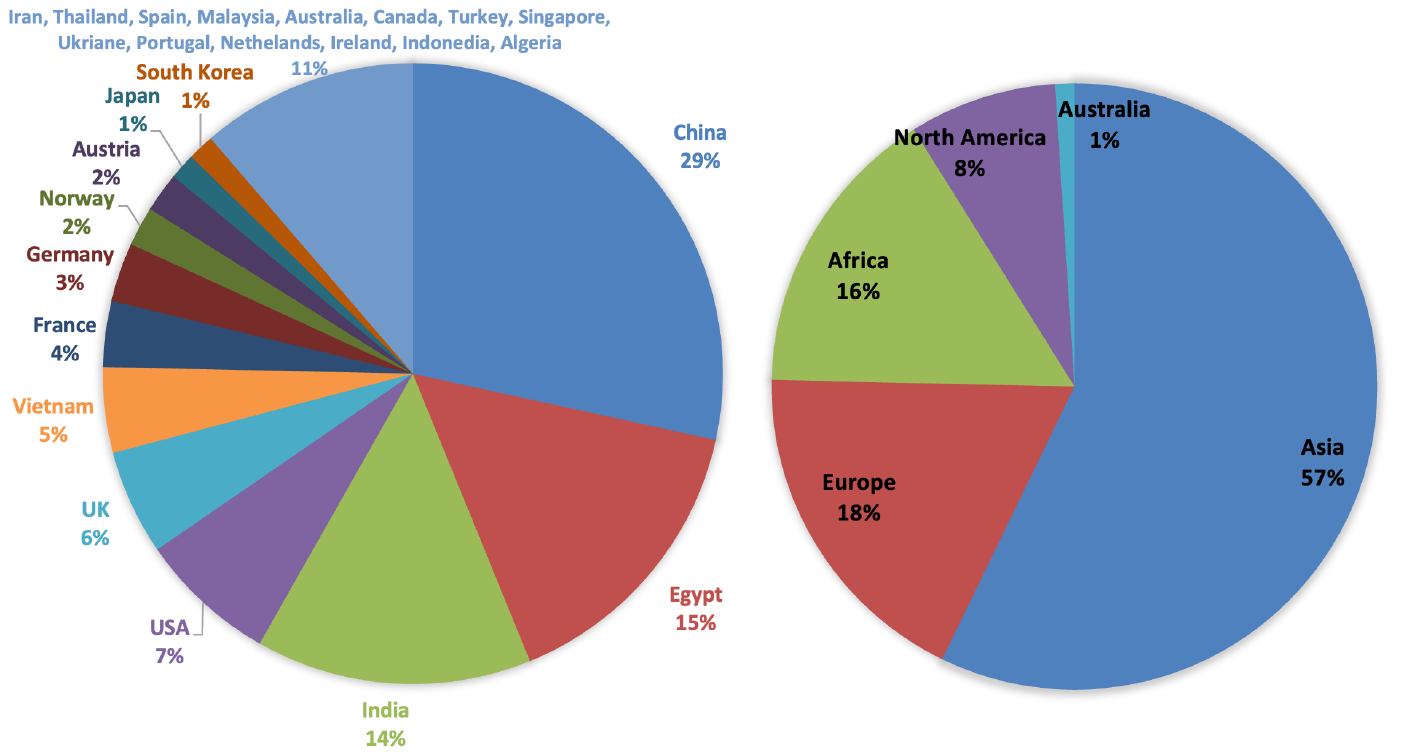}
    \caption{Geographical distribution of participants: (left) by country and (right) by continent.}
    \label{fig:pinechart}
\end{figure}

\vspace{3mm}
In total, \textbf{241} teams registered for WBCBench 2026. Fig.~\ref{fig:pinechart} shows the geographical distribution of participants, which are clearly spread across the globe, highlighting the importance of the WBC classification problem. Of these, \textbf{101} teams submitted valid final results. While top-performing methods substantially improve over our baselines, all approaches still struggle with blast and rare classes under heavy degradation. WBCBench 2026 therefore provides a rigorous, clinically grounded benchmark for robust WBC classification and highlights open challenges in deploying automated morphology analysis in routine haematology workflows.

This paper describes the dataset (\S\ref{sec:dataset}), degradation strategy (\S\ref{sec:degradation}), baseline methods and training protocol (\S\ref{sec:methods}), challenge evaluation criteria (\S\ref{sec:evaluation}), and results including baselines and participating teams (\S\ref{sec:results}). We conclude with a discussion of implications and open problems in robust WBC morphology analysis (\S\ref{sec:conclusion}).

\section{Existing datasets and challenges}

Several public WBC datasets (e.g.\ Raabin-WBC~\cite{raabin}, ALL-IDB~\cite{allidb} and related collections) have stimulated research in this area. Yet most exhibit modest sample sizes, coarse class taxonomies, or inadequate documentation of patient-level splits and acquisition conditions. Moreover, there is still no widely accepted benchmark that jointly targets:
(i) fine-grained morphology across many WBC classes,
(ii) severe and realistic class imbalance, and
(iii) robustness under controlled acquisition variability.

Benchmarking robustness to corruptions and domain shift has been widely explored in natural image classification (e.g.\ ImageNet-C and related suites). In medical imaging, analogous robustness benchmarks exist in some modalities, but not for haematological microscopy. WBCBench 2026 is designed to fill this gap by explicitly encoding patient-level separation, class imbalance and synthetic domain shift into a single, standardised evaluation.

\section{WBCBench 2026 Dataset}
\label{sec:dataset}

\subsection{Clinical context and annotation}

WBCBench 2026 comprises 55,012 microscopic images derived from 493 patients. All images in the dataset are microscopic peripheral blood smear patches acquired at a single institution using standardised Wright--Giemsa staining and a fixed imaging pipeline. Cells originate from patients routinely investigated for suspected haematologic disease, including leukaemia. The dataset covers 13 WBC classes (shown in Fig.~\ref{fig:examples}), including segmented neutrophils (SNE), band neutrophils (BNE), eosinophils (EO), basophils (BA), lymphocytes (LY), monocytes (MO), metamyelocytes (MMY), myelocytes (MY), promyelocytes (PMY), blasts (BL), variant lymphocytes (VLY), plasma cells (PC) and prolymphocytes (PLY).

Each image is a $368\times 368$ RGB patch centred on a single WBC. Class labels were assigned by 5 experienced haematopathologists, with ambiguous cases discussed in consensus sessions. Severe class imbalance is intrinsic to the clinical population. Mature neutrophils and lymphocytes account for most images, while blasts and other immature forms appear in only a small fraction of patients. Rather than artificially balancing classes, WBCBench preserves this distribution to reflect real diagnostic difficulty.

\subsection{Splits and phases}

\begin{figure*}[t]
    \centering
    \includegraphics[width=\textwidth]{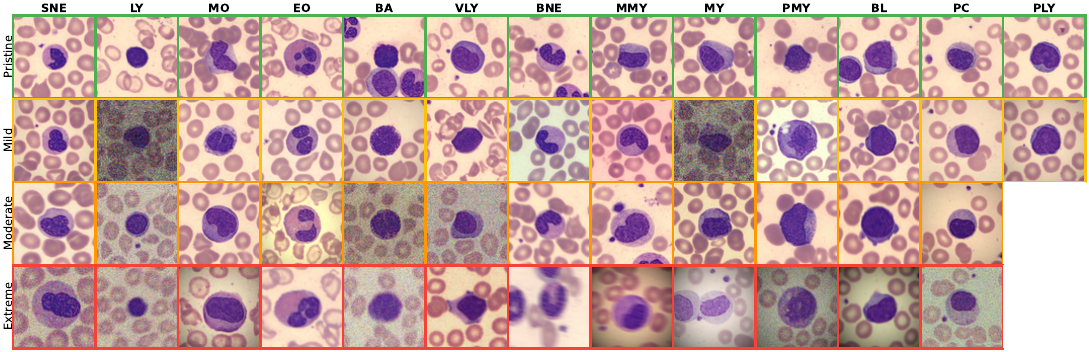}
    \caption{WBCBench 2026 representative images for all 13 WBC classes (columns) at four degradation severity levels (rows: Pristine, Mild, Moderate, Extreme). Coloured borders indicate severity. PLY moderate/extreme cells are absent due to rare-class protection (no moderate/extreme degradation applied to classes with ${<}0.5\%$ coverage).}
    \label{fig:examples}
    \vspace{-3mm}
\end{figure*}

The dataset is partitioned at the \emph{patient} level to prevent any image from the same patient appearing in both training and test, thereby avoiding information leakage. A total of $15\%$ of patients (74 patients, 8,288 images) are allocated to the Phase~1 training set, which is fully pristine. Phase~1 is released initially to allow participants to develop methods 
on pristine data with known class distributions. The remaining $85\%$ of patients are used to construct Phase~2, which introduces controlled synthetic degradation. Phase~2 splits are created using group-stratified sampling over patients to preserve coverage of minority classes, particularly blasts and other rare subtypes, in both validation and test. 
Details of the dataset partitioning are provided in Table~\ref{tab:deg_split}, and Figure~\ref{fig:examples} illustrates representative images from common and rare classes under pristine and degraded conditions.

\begin{table}[t]
\centering
\caption{Dataset partitioning and degradation severity fractions per split.}
\begin{tabular}{lcccc}
\hline
Split                                                                 & \begin{tabular}[c]{@{}c@{}}Phase 1 \\ train\end{tabular} & \begin{tabular}[c]{@{}c@{}}Phase 2 \\ train\end{tabular} & \begin{tabular}[c]{@{}c@{}}Phase 2 \\ eval\end{tabular} & \begin{tabular}[c]{@{}c@{}}Phase 2 \\ test\end{tabular} \\ \hline
\begin{tabular}[c]{@{}l@{}}Patients\\ Proportion\end{tabular}         & \begin{tabular}[c]{@{}c@{}}74\\ 15\%\end{tabular}          & \begin{tabular}[c]{@{}c@{}}222\\ 45\%\end{tabular}         & \begin{tabular}[c]{@{}c@{}}49\\ 10\%\end{tabular}         & \begin{tabular}[c]{@{}c@{}}148\\ 30\%\end{tabular}        \\
Num Images                                                            & 8\,288                                                     & 24\,897                                                    & 5\,350                                                    & 16\,477                                                   \\
Pristine                                                              & 1.00                                                       & 0.07                                                       & 0.40                                                      & 0.35                                                      \\
Mild                                                                  & 0.00                                                       & 0.55                                                       & 0.45                                                      & 0.45                                                      \\
Moderate                                                              & 0.00                                                       & 0.30                                                       & 0.12                                                      & 0.15                                                      \\
Extreme                                                               & 0.00                                                       & 0.08                                                       & 0.03                                                      & 0.05                                                      \\ \hline
\end{tabular}
\label{tab:deg_split}
\end{table}

\begin{table}[t]
\centering
\caption{Severity-dependent parameters. Parameters are sampled uniformly; motion blur angle ${\sim}\,U(0^{\circ},360^{\circ})$; $\Delta S$, $\Delta I$: additive offsets in HSV and pixel [0--255] scale, respectively.}
\resizebox{\columnwidth}{!}{
\setlength{\tabcolsep}{3pt}
\scriptsize
\begin{tabular}{llccc}
\toprule
Operator & Parameter & Mild & Moderate & Extreme \\
\midrule
\multicolumn{5}{c}{\textit{Geometric, blur \& noise}} \\
\midrule
Crop          & max ratio $r$              & 0.02--0.10 & 0.10--0.20 & 0.20--0.30 \\
Gaussian blur & $\sigma$ [px]              & 0.2--1.0   & 1.0--2.0   & 2.0--3.0   \\
Motion blur   & kernel $k$ [px]            & 3--15      & 5--25      & 20--35     \\
              & apply prob.\ $p_m$         & 0.1--0.3   & 0.3--0.7   & 0.7--1.0   \\
Gaussian noise& std.\ $\sigma_n$           & 2--8       & 8--18      & 18--25.5   \\
Poisson noise & rate $\lambda$             & 0.5--2.0   & 2.0--4.0   & 3.5--5.0   \\
\midrule
\multicolumn{5}{c}{\textit{Colour \& illumination}} \\
\midrule
Saturation    & $\Delta S$ (HSV)           & $[-10,10]$ & $[-15,15]$ & $[-25,25]$ \\
Gamma         & $\gamma$                   & 0.9--1.1   & 0.8--1.3   & 0.5--1.5   \\
Brightness    & $\Delta I$ [0--255]        & $[-10,10]$ & $[-15,15]$ & $[-25,25]$ \\
Vignetting    & strength $s$               & 0.1--0.3   & 0.3--0.6   & 0.5--0.8   \\
Colour jitter & factor $b$                 & 0.95--1.05 & 0.90--1.10 & 0.85--1.15 \\
              & prob.\ $p_{\text{cj}}$     & 0.5--1.0   & 0.7--1.0   & 0.8--1.0   \\
\bottomrule
\end{tabular}}

\label{tab:deg_params}
\end{table}
\subsection{Image Degradation Strategy}
\label{sec:degradation}

To emulate realistic acquisition artefacts and stress-test robustness, we apply a controlled synthetic degradation pipeline to Phase~2 data only. All Phase~1 images remain pristine.

\subsubsection{Severity assignment} Phase~2 images are degraded to four severity levels (pristine, mild, moderate, extreme) with specific splits in Table~\ref{tab:deg_split}, yielding ${\approx}30\%$ pristine in the combined training set. 
For PLY, the default pristine rate would round to zero, leaving no clean morphological reference; we therefore enforce the propotion $p_{\text{pris}}\!=\!0.70$, $p_{\text{mild}}\!=\!0.30$, $p_{\text{mod}}\!=\!p_{\text{ext}}\!=\!0$, and a fallback rule guarantees at least one undegraded image for every class.

\subsubsection{Degradation operators} For each non-pristine image, operators 
are randomly selected and applied sequentially: 1--3 for mild, 1--4 for 
moderate, and 3--5 for extreme. Crucially, operator \emph{parameters} 
are drawn from progressively stronger ranges per severity level 
(Table~\ref{tab:deg_params}), so severity is determined by both the 
number of operators and their intensity. JPEG compression and 
additional rotations are disabled to avoid unrealistic artefacts.

\section{Baselines and Evaluation}
\label{sec:methods}
\begin{enumerate}
\item [\textbf{A.}] \textbf{Baseline Models}

The challenge does not prescribe a specific modelling approach; participants are free to design arbitrary architectures and training strategies. To provide a reference point, we implement two baselines~\cite{clifton2025mamba}.


\begin{itemize}
    \item \textbf{Convolutional networks.} ResNet-50~\cite{he2016deep} initialised from ImageNet pretraining, fine-tuned end-to-end with a 13-way linear classifier.

    \item \textbf{Vision Transformers.} Swin-Tiny~\cite{swin} adapted to $224\times 224$ inputs with a classification head, initialised from ImageNet-pretrained weights.

\end{itemize}

\noindent Baseline methods are trained with cross-entropy loss weighted inversely by class frequency, using AdamW with cosine learning rate decay and label smoothing ($\epsilon = 0.05$).


\label{sec:evaluation}

    \item[\textbf{B.}] \textbf{Metrics }

The primary ranking metric is the \emph{macro-averaged F1 score} computed over 13 WBC classes on \texttt{phase2\_test}:
\[
\mathrm{F1}_{\mathrm{macro}} = \frac{1}{13} \sum_{c=1}^{13} \frac{2\,P_c R_c}{P_c + R_c},
\]
where $P_c$ and $R_c$ denote the precision and recall for class $c$, respectively. Macro-F1 is deliberately insensitive to class frequency, penalising models that ignore rare cells even when overall accuracy is high. Tie-breaking uses balanced accuracy, then macro precision, then macro specificity, in that order.

\end{enumerate}

\section{Results and Discussion}
\label{sec:results}

A total of \textbf{241} teams registered for WBCBench 2026, spanning academia, industry and independent researchers, among which \textbf{101} teams submitted at least one valid set of predictions~\cite{srivastava2026ensemble, gitau2026multistage,xiao2026foundation,nguyen2026synergizing,le2026robust,ng2026hierarchical}. Among the participants, \textbf{73} (72\%) exceeded the ResNet-50 baseline ($0.635$) and \textbf{66} (65\%) surpassed the stronger Swin-Tiny baseline ($0.643$). \textbf{7} teams achieved macro-F1 above $0.70$ and \textbf{2} above $0.75$. The median macro-F1 across all participants was $0.656$ (mean $0.629$), indicating that top-ranked methods outperform the typical submission by a substantial margin. The top-10 results are reported in Table~\ref{tab:leaderboard}.

\begin{table}[t]
    \centering
    \resizebox{\columnwidth}{!}{
    \setlength{\tabcolsep}{3.2pt}
    \scriptsize
    \begin{tabular}{rlcccc}
        \toprule
        Rank & Team & Macro-F1 & Bal.\ Acc & Macro Prec & Macro Spec \\
        \midrule
        1  & FDVTS\_WBC~\cite{xiao2026foundation}     & \textbf{0.777} & \textbf{0.753} & 0.818 & \textbf{0.996} \\
        2  & PathMedAI~\cite{nguyen2026synergizing}        & 0.771 & 0.733 & \textbf{0.834} & 0.994 \\
        3  & jht010312        & 0.740 & 0.742 & 0.756 & 0.995 \\
        4  & CPRL             & 0.720 & 0.701 & 0.759 & 0.995 \\
        5  & PACV             & 0.719 & 0.707 & 0.746 & 0.995 \\
        6  & AIO-MHIL\cite{le2026robust}         & 0.708 & 0.693 & 0.738 & 0.995 \\
        7  & Quan H.\ Cap     & 0.704 & 0.719 & 0.695 & 0.995 \\
        8  & GODA             & 0.686 & 0.670 & 0.710 & 0.995 \\
        9  & jingxin2001      & 0.684 & 0.659 & 0.728 & 0.994 \\
        10 & smart\_lab\cite{ng2026hierarchical}       & 0.682 & 0.681 & 0.688 & \textbf{0.996} \\
        \midrule
        -- & ResNet-50~\cite{he2016deep} & 0.635 & 0.624 & 0.653 & 0.994 \\
        -- & Swin-Tiny~\cite{swin}       & 0.643 & 0.629 & 0.663 & 0.994 \\
        \bottomrule
    \end{tabular}}
    \caption{Final leaderboard: top-10 results on \texttt{phase2\_test} with organiser baselines.}
    \label{tab:leaderboard}
    \vspace{-2mm}
\end{table}

\subsection{Participant methods and code review}

Top-ranked participants submitted code for independent verification. We summarise the five submitted methods below.

\textbf{FDVTS\_WBC~\cite{xiao2026foundation} (Rank 1, macro-F1\,=\,0.777).} Hierarchical ensemble of three foundation models---DinoBloom-B, DINOv3-ConvNeXt-Base, and DINOv3-ViT-B/16---plus a dedicated binary classifier for the rare PLY class combining patch-level voting and exemplar matching. Self-training via pseudo-labelling of unlabelled test images is used to improve generalisation. Focal loss with effective-number class weighting is applied.

\textbf{PathMedAI~\cite{nguyen2026synergizing} (Rank 2, macro-F1\,=\,0.771).} Five-fold ensemble of Swin-Small Transformers augmented with a MedSigLIP contrastive embedding head for rare-class separation. A Pix2Pix GAN denoising step is applied to images at inference. Custom geometric filters (``spikiness score'' for PLY, Mahalanobis OOD filter for PC) and per-class sensitivity calibration further refine predictions.

\textbf{jht010312 (Rank 3, macro-F1\,=\,0.740).} DinoBloom fine-tuned with LoRA (rank\,=\,64) and an ETF classifier head. Cellpose-based cell denoising is applied during preprocessing. LogitAdjustmentLoss corrects majority-class bias; Mixup augmentation is used during training. External labelled datasets (APL/AML, BCCD, CellWiki) were integrated.

\textbf{AIO-MHIL~\cite{le2026robust} (Rank 6, macro-F1\,=\,0.708).} Two-stage decoupled training: (1) full ResNet-50/ResNet-152 network trained with cross-entropy loss and random sampling; (2) frozen backbone with classifier head retrained using class-balanced sampling and a combined cross-entropy + focal loss. Macenko stain normalisation is applied at preprocessing. Inference uses 8$\times$ test-time augmentation and weighted probability averaging.

\textbf{smart\_lab~\cite{ng2026hierarchical} (Rank 10, macro-F1\,=\,0.682).} DINOv2\slash DinoBloom backbone fine-tuned with LoRA (rank\,=\,128/64 across splits) across multiple data splits. Inference employs a hierarchical coarse-to-fine strategy combining kNN retrieval and proxy-based NCA++ at three semantic levels, with final predictions by majority voting across 3 splits $\times$ 5 folds. External public datasets (PBC\_dataset\_normal\_DIB, Raabin-WBC) were used for training.

\subsection{Per-class difficulty analysis}

Table~\ref{tab:perclass} reports average per-class F1 across all ten top-ranked submissions. Mature, high-support classes (SNE, LY, EO, BA) are well-classified by all participants, with mean F1 above 0.97. The most challenging classes are \textbf{PC}, \textbf{BNE} , \textbf{PMY}, and \textbf{PLY}. PC ($n=15$, mean F1\,=\,0.15) is severely underrepresented, and most models fail to predict any instances, resulting in near-zero F1 scores. PLY ($n=2$, mean F1\,=\,0.59) is similarly difficult to assess due to the extremely limited number of test samples. BNE ($n=138$, mean F1\,=\,0.42) represents a morphologically transitional stage and is frequently misclassified as SNE or MMY. PMY ($n=48$, mean F1\,=\,0.50) poses additional difficulty due to both morphological overlap with adjacent developmental stages and limited representation. These findings highlight that imbalance and morphological ambiguity, rather than degradation alone, are the primary bottlenecks.

Crucially, high micro-averaged accuracy can coexist with poor macro-F1 when rare classes are ignored. The challenge design deliberately exposes this failure mode: top-performing methods succeed by boosting recall on PMY, BNE, and PC without sacrificing performance on abundant classes.

\begin{table}[t]
    \centering
    \resizebox{0.65\columnwidth}{!}{
    \setlength{\tabcolsep}{3.5pt}
    \scriptsize
    \begin{tabular}{lrrrr}
        \toprule
        Class & Support & Avg F1 & Min F1 & Max F1 \\
        \midrule
        SNE   & 8188 & 0.984 & 0.981 & 0.987 \\
        LY    & 4269 & 0.960 & 0.953 & 0.964 \\
        MO    & 1508 & 0.923 & 0.902 & 0.936 \\
        EO    &  492 & 0.968 & 0.949 & 0.979 \\
        BA    &  303 & 0.969 & 0.947 & 0.985 \\
        BL    &  902 & 0.871 & 0.843 & 0.892 \\
        MY    &  184 & 0.752 & 0.670 & 0.784 \\
        VLY   &  253 & 0.625 & 0.570 & 0.673 \\
        MMY   &  175 & 0.637 & 0.566 & 0.701 \\
        PLY   &    2 & 0.590 & 0.000 & 1.000 \\
        PMY   &   48 & 0.498 & 0.430 & 0.562 \\
        BNE   &  138 & 0.423 & 0.319 & 0.465 \\
        PC    &   15 & 0.150 & 0.000 & 0.800 \\
        
        \bottomrule
    \end{tabular}}
    \caption{Per-class F1 averaged over the top-10 submissions, sorted by difficulty (easiest first). PLY and PC results are statistically unreliable due to very low test support.}
    \label{tab:perclass}
    \vspace{-2mm}
\end{table}

\section{Conclusion}
\label{sec:conclusion}

We presented WBCBench 2026, an ISBI challenge and benchmark targeting robust white blood cell classification under realistic class imbalance and synthetic domain shift. The dataset comprises single-site, expert-annotated blood smear images spanning 13 WBC classes, including blasts and other rare subtypes. With 241 registered teams and 101 valid submissions, the challenge demonstrates strong community interest. Top submissions based on domain-adaptive foundation models and specialised rare-class pipelines outperform vanilla baselines by over 13 macro-F1 points, yet blasts and rare cells remain challenging under heavy degradation. The dataset, evaluator, and baseline implementations are publicly released to support continued research in robust haematological image analysis.

\section{COMPLIANCE WITH ETHICAL STANDARDS}

This study was performed in line with the principles of the Declaration of Helsinki. The dataset was commercially obtained from Chulalongkorn University. Additional ethical approval was not required, as confirmed by the license.

{\small
\bibliographystyle{IEEEbib}
\bibliography{refs}
}

\end{document}